\documentclass[runningheads]{llncs}
\usepackage[T1]{fontenc}
\usepackage{graphicx}

\usepackage{makecell}
\usepackage{bbding}
\usepackage{bm}

\newcommand{\scombined}{Ensemble-BC+S}
\newcommand{\ethree}{Ensemble-S}
\newcommand{\vote}{Ensemble-V}
\newcommand{\esix}{Ensemble-S$\rm ^{-S}$}
\newcommand{\eone}{Ensemble-S$\rm _A$}
\newcommand{\efive}{Ensemble-S$\rm _B$}
\newcommand{\efour}{Ensemble-S$\rm _{B}^{-GS}$}
\newcommand{\eseven}{Ensemble-S$\rm ^{RGB}$}
\newcommand{\bc}{BC}
\newcommand{\rgb}{Skinny-RGB}
\newcommand{\gs}{Skinny-GS}
\newcommand{\skin}{Skinny-$\rm P_B^S$}
\newcommand{\nonskin}{Skinny-$\rm P_B^{NS}$}

\newcommand{\skinreg}{$\rm P_B^S$}
\newcommand{\nonskinreg}{$\rm P_B^{NS}$}

\begin{document}
\title{Ensembling convolutional neural networks for human skin segmentation}



%
%

\author{Patryk Kuban\inst{1} and\\
Michal Kawulok\inst{1}\orcidID{0000-0002-3669-5110}}
\institute{Department of Algorithmics and Software\\ Silesian University of Technology, Gliwice, Poland \\ \email{michal.kawulok@polsl.pl}}

\authorrunning{P. Kuban and M. Kawulok}
%
\maketitle              
\begin{abstract}
Detecting and segmenting human skin regions in digital images is an intensively explored topic of computer vision with a variety of approaches proposed over the years that have been found useful in numerous practical applications. The first methods were based on pixel-wise skin color modeling and they were later enhanced with context-based analysis to include the textural and geometrical features, recently extracted using deep convolutional neural networks. It has been also demonstrated that skin regions can be segmented from grayscale images without using color information at all. However, the possibility to combine these two sources of information has not been explored so far and we address this research gap with the contribution reported in this paper. We propose to train a convolutional network using the datasets focused on different features to create an ensemble whose individual outcomes are effectively combined using yet another convolutional network trained to produce the final segmentation map. The experimental results clearly indicate that the proposed approach outperforms the basic classifiers, as well as an ensemble based on the voting scheme. We expect that this study will help in developing new ensemble-based techniques that will improve the performance of semantic segmentation systems, reaching beyond the problem of detecting human skin.

\end{abstract}
\section{Introduction} \label{sec:intro}

Human skin detection consists in taking a binary decision on whether the skin is present in digital imagery data at different granularity levels (a pixel, an image region, a whole image, or a video sequence)~\cite{Lumini2020}. Commonly, it is performed at a pixel level and therefore consists in segmenting skin regions~\cite{Tarasiewicz2020}. Skin segmentation plays a pivotal role in gesture recognition~\cite{Rahim2020}, objectionable content filtering~\cite{Lee2007}, privacy protection~\cite{Shifa2020}, selective image compression~\cite{Rodrigues2006}, skin rendering~\cite{Poirier2004}, and more~\cite{Senouci2017}. The recent techniques underpinned with deep learning allow for extracting skin regions from color~\cite{Tarasiewicz2020,Ma2018} and grayscale~\cite{Paracchini2020,XuSarkar2022} images, however it remains a challenging problem of computer vision that requires a thorough visual scene understanding, as many other problems related with semantic segmentation~\cite{MuhammadHussain2022}.

\subsection{Related Work}

The first attempts to detect skin in digital images relied on modeling the skin color in a variety of color spaces~\cite{Naji2019}. By applying a set of handcrafted rules~\cite{Chen2012} or a learned model~\cite{Jones2002}, every pixel can be classified as skin or non-skin based on its position in the color space. The large amount of pixels in the training set coupled with low dimensionality of color spaces (commonly limited to three dimensions) made the Bayesian classifier (BC) highly effective in this case~\cite{Jones2002,Phung2005}. However, as skin appearance varies across different individuals and image acquisition conditions, its color alone is not a sufficiently discriminating feature for segmenting the skin regions effectively. Therefore, features extracted from a wider context, capturing the texture or geometrical properties, allow for enhancing the segmentation performance~\cite{Kawulok2014PRL}.

As demonstrated in an extensive experimental study by Lumini et al.~\cite{Lumini2020}, the handcrafted features have been surpassed by those learned with convolutional neural networks (CNNs). A network-in-network (NiN) architecture introduced by Kim et al.~\cite{Kim2017} outperformed the BC and the techniques based on handcrafted textural features, as well as the VGG network~\cite{Simonyan2014} trained for segmenting skin areas. Zuo et al. proposed to combine the CNN with a recurrent network~\cite{Zuo2017}, and Arsanal et al. demonstrated the benefits of applying residual connections for this purpose~\cite{Arsalan2020}. These methods were subsequently outperformed with the Skinny network introduced by Tarasiewicz et al.~\cite{Tarasiewicz2020}, which is a lightweight U-Net architecture with inception modules and dense blocks. The segmentation outcome can also benefit from appropriate postprocessing~\cite{Baldissera2021}, including the use of morphological filters~\cite{Lumini2019}. In~\cite{DingLiu2023}, color attention mechanism was proposed to decrease the computational burden and improve the real-time performance.

There were also some efforts aimed at improving the training process, including coupling body and skin detection in a semi-supervised fashion~\cite{He2019} and exploring data augmentation techniques to improve the robustness against varying lighting conditions~\cite{YouLee2023}. Appropriate data augmentation performed in the spectral dimension may also help in learning color invariant features~\cite{XuSarkar2022}. Another possibility explored in the literature is to rely exclusively on grayscale images---this was motivated by the fact that human observers are capable to identify the skin regions using textural and context information~\cite{Sarkar2017}. Paracchini et al. proposed to train an encoder-decoder architecture with grayscale images, thus making the segmentation fully independent from color information. The grayscale image does not contain any additional information compared with the color one, so the networks trained from color images may also exploit the features present in grayscale images. However, the color-based features are definitely easier to extract and without appropriate guidance, the geometrical and texture features present in the grayscale image may not be fully exploited during training. While several image segmentation networks can be treated as an ensemble whose responses are averaged to improve the final segmentation~\cite{Nanni2023}, to our best knowledge it has not been attempted to combine networks trained from grayscale and color information.

\subsection{Contribution}

In the research reported here, we address the identified research gap related with extracting and coupling different types of skin-presence features. In particular, our contribution can be summarized in the following points.
\begin{enumerate}
    \item We propose a new approach towards constructing ensembles of homogeneous image segmentation networks trained in different ways that are combined in a sequential manner to improve the segmentation outcome. We demonstrate that this approach is more effective than voting-based ensembles.
    \item We exploit the simple BC-based skin segmentation for training multiple CNNs that are later treated as an ensemble.
    \item We demonstrate that the  CNNs trained  from grayscale and color images are focused on different features, thus they can be effectively combined within the proposed ensemble scheme.
\end{enumerate}

\section{Proposed Approach} \label{sec:method}

The proposed method consists in preparing a set of diverse deep CNN models, whose outcomes are fed to the second-level CNN that integrates the first-level decisions. In Section~\ref{sec:baseline}, we outline the exploited baseline techniques, namely the Skinny CNN~\cite{Tarasiewicz2020} that serves as a base model in our ensemble and BC-based skin segmentation~\cite{Phung2005} which we employ to increase the diversity of the trained models. The details of our approach to constructing skin segmentation ensembles is presented in Section~\ref{sec:ensemble}.

\begin{figure}[b]
    \centering
    \includegraphics[width=\textwidth]{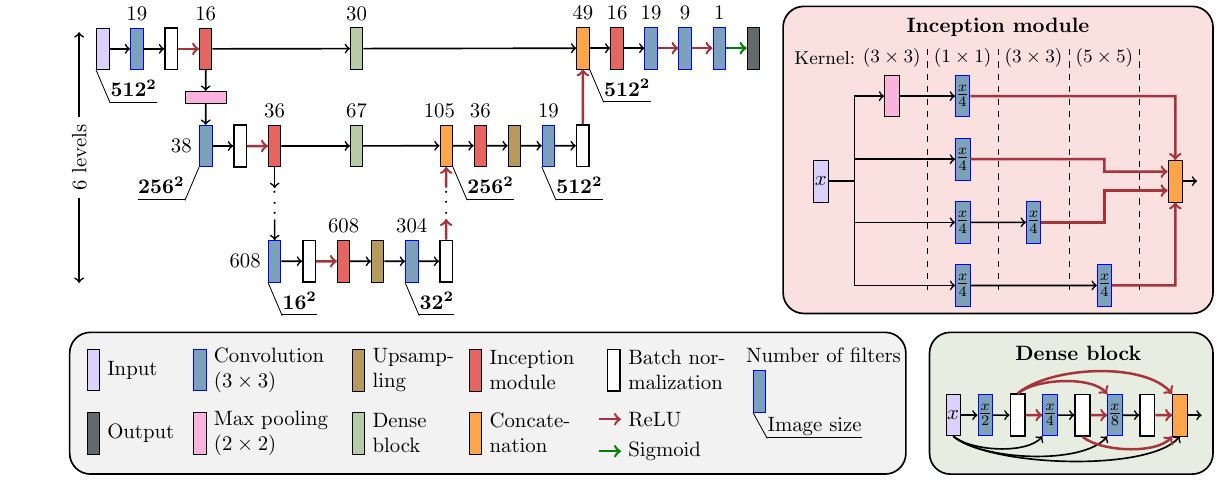}
    \caption{Architecture of the exploited Skinny network for human skin segmentation.}
    \label{fig:skinny}
\end{figure}
\subsection{Baseline Techniques} \label{sec:baseline}
We have selected the Skinny network~\cite{Tarasiewicz2020} to construct our CNN ensemble, as it is easy to train while offering the state-of-the-art performance (Fig.~\ref{fig:skinny}). It is a lightweight U-Net architecture~\cite{DBLP:journals/corr/RonnebergerFB15} composed of six levels and $7.5\cdot 10^6$ learnable parameters. We used Skinny both to obtain the base classifiers, as well as for aggregating the decisions to obtain the final segmentation result. By default, it processes  color images with red, green and blue (RGB) channels, but it can be straightforwardly applied to processing data of different modality, including single-channel grayscale images.

To employ the BC classifier for skin segmentation~\cite{Phung2005}, at first we compute the histograms for the skin ($C_s$)
and non-skin ($C_{ns}$) classes and we obtain the probability of
observing a given color value ($v$) in the $C_x$ class as: $P(v|C_x) = C_x(v)/N_x$,
where $C_x(v)$ is the number of pixels in the $x$-th class having $v$ color
and $N_x$ is the number of pixels in that class. The
probability that a pixel of $v$ color presents skin is then obtained as:
$    P(C_s|v) = \frac
                    {P(v|C_s)P(C_s)}
                    {P(v|C_s)P(C_s)+P(v|C_{ns})P(C_{ns})}$. The
\emph{a priori} probabilities are commonly set to $P(C_s) = P(C_{ns}) = 0.5$.

\subsection{Ensembles for Skin Segmentation} \label{sec:ensemble}

\begin{figure}[b]
    \centering
    \includegraphics[width=\textwidth]{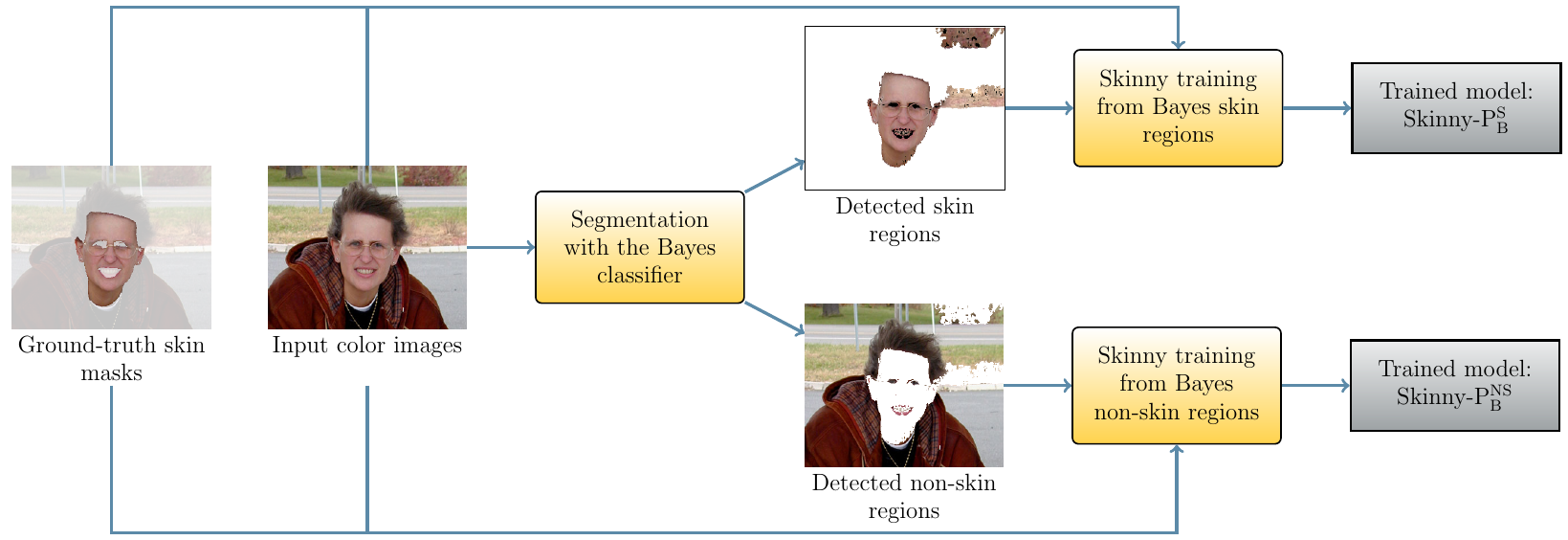}
    \caption{The BC-based skin segmentation employed to split the input images into skin and non-skin regions, from which two different Skinny models are trained.}
    \label{fig:double_branch}
\end{figure}
The proposed ensemble is composed of two levels---at the first one, we prepare a set of Skinny CNNs trained to focus on different features, which are combined using another Skinny CNN trained to aggregate the first-level responses into the final segmentation map. At first, we prepared two different base models trained from color and grayscale images, thus obtaining \rgb\, and \gs\, models, respectively. To further diversify the base models, we exploit the BC-based skin segmentation, as depicted in Fig.~\ref{fig:double_branch}. The BC splits every color image into skin (\skinreg) and non-skin (\nonskinreg) regions, and we train two Skinny models from these two exclusive training sets (\skin\, and \nonskin, respectively). During training, the Skinny is presented with the whole color image, but the loss function is computed only in the areas indicated as skin and non-skin by the BC for \skin\, and \nonskin, respectively. In this way, \skin\, is trained to eliminate the BC's false positives, while \nonskin\, is focused on refining false negative pixels. As the color feature is already exploited by the BC, we may expect that these two models are more focused on extracting other types of features. As the datasets obtained after BC-based stratification are imbalanced (especially \nonskin), for the loss function we couple the binary cross entropy with the Dice coefficient during training.

\begin{figure}
    \centering
    \includegraphics[width=\textwidth]{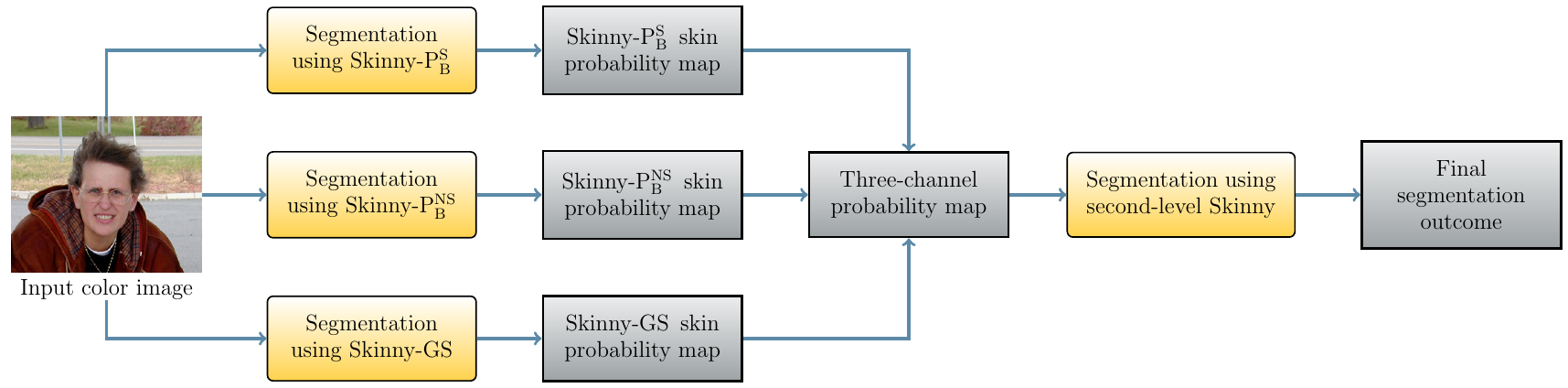}
    \caption{A flowchart presenting skin segmentation using \ethree. The probability maps retrieved with three Skinny models trained from the grayscale image and from two exclusive regions in the color image, indicated as skin and non-skin by the BC, are aggregated using the second-level Skinny model to obtain the final segmentation outcome.}
    \label{fig:ensemble}
\end{figure}

The final ensemble (\ethree) is outlined in Fig.~\ref{fig:ensemble}. Three Skinny models trained from the grayscale images and from the regions indicated as skin and non-skin by the BC are employed to retrieve three skin-presence probability maps. These maps are stacked together to form an input (three-channel image) that is presented to the second-level Skinny which renders the final segmentation outcome. During our research, we have considered different variants of ensembling a variety of Skinny models which are discussed later in Section~\ref{sec:exp}.

\section{Experimental Validation} \label{sec:exp}
\begin{table}[b!]
\renewcommand{\tabcolsep}{2mm}
    \caption{The investigated base models for skin segmentation.}
    \centering
    \begin{tabular}{lp{8.5cm}}
    \Xhline{2\arrayrulewidth}
    \textbf{Method name} &  \textbf{Description} \\
    \Xhline{2\arrayrulewidth}
         \bc & Bayesian classifier trained over the RGB color space \\
         \rgb & Skinny network trained over the RGB color space (a baseline Skinny model) \\
         \gs & Skinny trained from the grayscale images (single-channel) \\
         \skin & Skinny trained over the regions indicated as skin by the BC \\
         \nonskin & Skinny trained over the regions indicated as non-skin by BC \\

    \Xhline{2\arrayrulewidth}
    \end{tabular}
    \label{tab:base}
\end{table}
We have validated the proposed approach using the ECU skin dataset~\cite{Phung2005} that contains 4000 color images which we have randomly split into training, validation and test sets with 1750, 250 and 2000 images, respectively. To decrease the computational burden, we downsampled all the images preserving the aspect ratio, so that they do not exceed the size of $256\times 256$ pixels. The Skinny network was trained in 200 epochs with the learning rate set to $10^{-3}$. For the quantitative evaluation, we report the precision, recall and F-score metrics, and we also render the precision-recall curves. The investigated base models are enlisted in Table~\ref{tab:base} and the constructed ensembles are shown in Table~\ref{tab:ensembles}. Skinny requires around 50\,ms when run using NVIDIA RTX 2080Ti (11\,GB VRAM) GPU to process a single image.
\begin{table}[t!]
\renewcommand{\tabcolsep}{2mm}
    \caption{The constructed ensembles investigated in our experimental study that aggregate the grayscale channel (GS) and skin probability maps extracted using different Skinny models.}
    \centering
    \resizebox{\textwidth}{!}{
    \begin{tabular}{lllccccc}
    \Xhline{2\arrayrulewidth}
    & \textbf{Method } & \textbf{Ensembling } & \multicolumn{5}{c}{\textbf{Input channel}} \\ \cline{4-8} \\[-3.5mm]
    & \textbf{name} & \textbf{technique} & GS & \rgb & \gs & \skin & \nonskin \\
    \Xhline{2\arrayrulewidth}

        & \ethree & Skinny & & & \Checkmark & \Checkmark & \Checkmark \\
        & \vote & Voting &  & & \Checkmark & \Checkmark & \Checkmark \\
        & \eseven & Skinny & & \Checkmark & & \Checkmark & \Checkmark \\
        & \esix & Skinny &   & & \Checkmark &  & \Checkmark \\
        & \eone & Skinny & \Checkmark & & & \Checkmark & \Checkmark \\
        & \efive & Skinny & \Checkmark & \Checkmark & \Checkmark \\
        & \efour & Skinny &  & \Checkmark & \Checkmark \\
        & \scombined & BC selection & & & & \Checkmark & \Checkmark \\

    \Xhline{2\arrayrulewidth}
    \end{tabular}
    }
    \label{tab:ensembles}
\end{table}
\begin{table}[t]
    \centering
    \renewcommand{\tabcolsep}{4mm}
    \caption{Quantitative scores obtained with the base classifiers and with the investigated ensembles for the ECU test set. The best scores in each group are boldfaced. }
    \begin{tabular}{lccc}
       \Xhline{2\arrayrulewidth}
       \textbf{Method} & \textbf{F-score} & \textbf{Precision} & \textbf{Recall} \\
       \Xhline{2\arrayrulewidth}
\bc	& $0.7394$	& $0.7504$	& $0.7288$ \\
\rgb	& $0.8751$	& $0.8767$	& $0.8752$ \\
\gs	& $0.8020$	& $0.8146$	& $0.7899$ \\
\skin	& $0.8690$	& $0.8690$	& $0.8690$ \\
\nonskin	& $\bm{0.8789}$	& $\bm{0.8797}$	& $\bm{0.8781}$ \\ \hline
\ethree	& $\bm{0.8965}$	& $0.9001$	& $\bm{0.8930}$ \\
\vote	& $0.8718$	& $\bm{0.9142}$	& $0.8331$ \\
\eseven	& $0.8912$	& $0.8963$	& $0.8860$ \\	
\esix	& $0.8913$	& $0.8931$	& $0.8894$ \\
\eone	& $0.8893$	& $0.8883$	& $0.8903$ \\
\efive	& $0.8917$	& $0.8933$	& $0.8902$ \\
\efour	& $0.8864$	& $0.8900$	& $0.8883$ \\

\scombined	& $0.8757$	& $0.8749$	& $0.8764$ \\
\Xhline{2\arrayrulewidth}
    \end{tabular}
    \label{tab:scores}
\end{table}

\begin{figure}[t]
    \centering
    \includegraphics[width=\textwidth]{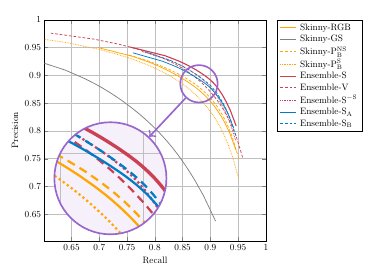}
    \caption{Precision-recall curves for the selected base models and ensemble classifiers for the ECU test set.}
    \label{fig:pc_curves}
\end{figure}

The obtained quantitative scores are reported in Table~\ref{tab:scores} and the precision-recall curves for the selected models are presented in Fig.~\ref{fig:pc_curves}. All the Skinny models, including \gs\, that does not exploit color information, render better scores than the BC, and the best result is retrieved with the model trained from the BC's non-skin regions (\nonskin). In order to justify our choice of the ensembling technique, we have considered several different variants of combining multiple segmentation outcomes (Table~\ref{tab:ensembles}). In \vote, we combine the outcomes retrieved with the same base models relying on the majority voting instead of employing the second-level Skinny model. Although the voting increases the precision compared with \ethree, the obtained recall is much worse, resulting in the F-score lower than for the base models. We also substituted the \gs\, model with \rgb\, in \eseven, and we tried excluding the weakest \skin\, base model (\esix)---in both cases, the results were worse than for \ethree. In \eone, we used the grayscale image instead of the \gs\, outcome and in \efive, we excluded the models trained from the data stratified using the BC (also without using the grayscale channel in \efour). Finally, in \scombined, we apply a two-branch approach, in which we select either \skin\, or \nonskin\, model for the pixels classified as skin or non-skin by the BC, respectively. All the ensembles that exploit Skinny at the second level are better than the best base model, with \ethree\, rendering the highest F-score. Although the quantitative differences between \ethree\, and other ensembles are not large, they are all statistically significant according the the two-tailed Wilcoxon test ($p<0.05$). Also, the precision-recall curve for \ethree\, dominates all the remaining techniques.

\begin{figure}
\newcommand{\mywidth}{0.145}
    \centering
    \scriptsize
    \resizebox{\textwidth}{!}{
    \begin{tabular}{lcccccc}
    & \rgb & \nonskin & \gs & \ethree & \vote & Ens.-BC+S \\

    \raisebox{5mm}{(a)} &
    \includegraphics[width=\mywidth\textwidth]{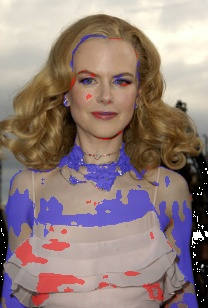} &
    \includegraphics[width=\mywidth\textwidth]{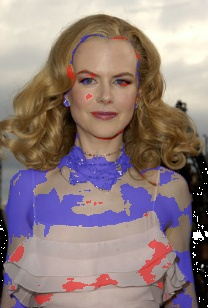} &
    \includegraphics[width=\mywidth\textwidth]{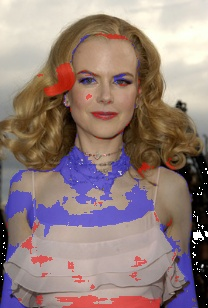} &
    \includegraphics[width=\mywidth\textwidth]{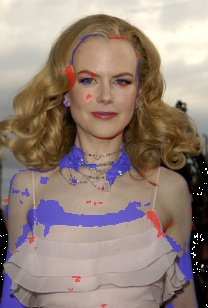} &
    \includegraphics[width=\mywidth\textwidth]{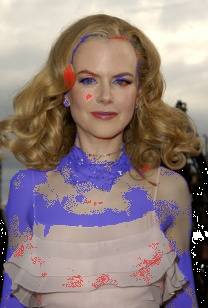} &
    \includegraphics[width=\mywidth\textwidth]{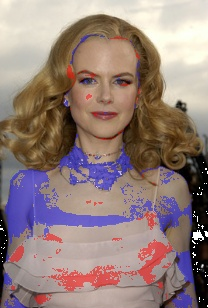} \\
& 0.7316	& 0.6978	& 0.7287	& \textbf{0.8581}	& 0.7274	& 0.7392\\

    \raisebox{5mm}{(b)} &
     \includegraphics[width=\mywidth\textwidth]{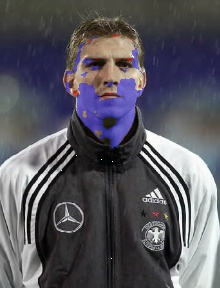} &
     \includegraphics[width=\mywidth\textwidth]{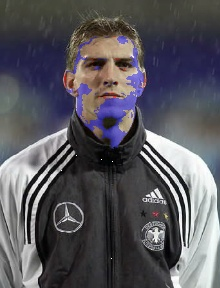} &
    \includegraphics[width=\mywidth\textwidth]{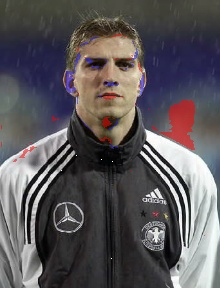} &
    \includegraphics[width=\mywidth\textwidth]{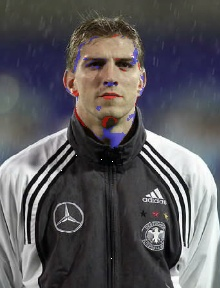} &
    \includegraphics[width=\mywidth\textwidth]{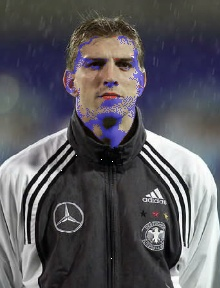} &
    \includegraphics[width=\mywidth\textwidth]{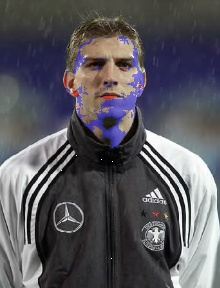} \\

& 0.6345	& 0.7617	& 0.8452	& \textbf{0.9238}	& 0.8101	& 0.7830\\

    \raisebox{5mm}{(c)} &
     \includegraphics[width=\mywidth\textwidth]{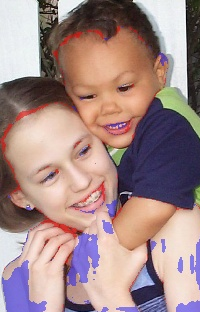} &
     \includegraphics[width=\mywidth\textwidth]{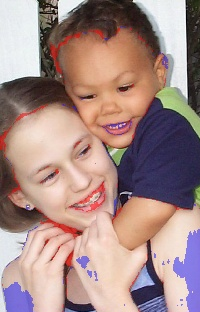} &
    \includegraphics[width=\mywidth\textwidth]{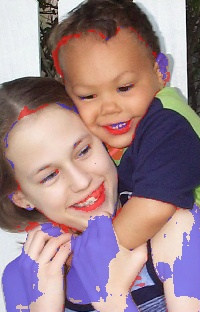} &
    \includegraphics[width=\mywidth\textwidth]{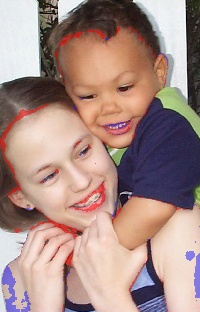} &
    \includegraphics[width=\mywidth\textwidth]{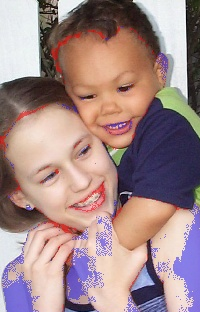} &
    \includegraphics[width=\mywidth\textwidth]{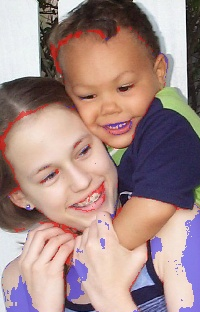} \\
    & 0.9132	& 0.9329	& 0.8404	& \textbf{0.9572}	& 0.9001	& 0.9197\\

    \raisebox{5mm}{(d)} &
     \includegraphics[width=\mywidth\textwidth]{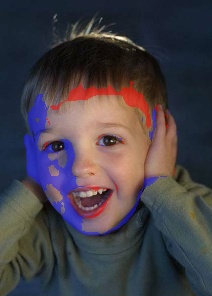} &
     \includegraphics[width=\mywidth\textwidth]{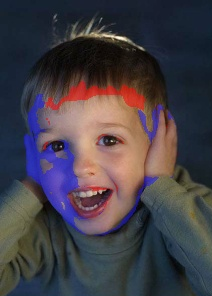} &
    \includegraphics[width=\mywidth\textwidth]{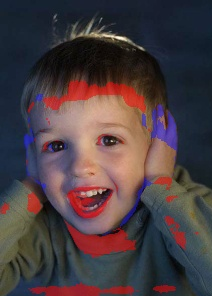} &
    \includegraphics[width=\mywidth\textwidth]{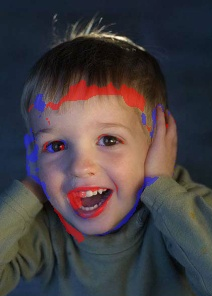} &
    \includegraphics[width=\mywidth\textwidth]{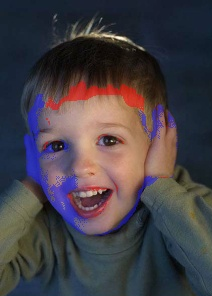} &
    \includegraphics[width=\mywidth\textwidth]{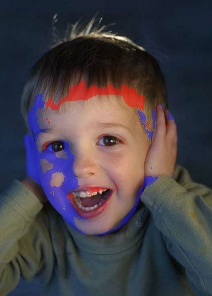} \\
& 0.8251	& 0.8180	& 0.8032	& \textbf{0.8965}	& 0.8497	& 0.8285\\

    \raisebox{5mm}{(e)} &
     \includegraphics[width=\mywidth\textwidth]{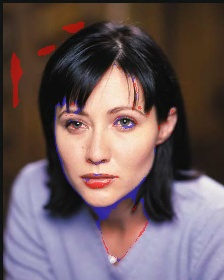} &
     \includegraphics[width=\mywidth\textwidth]{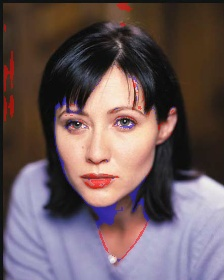} &
    \includegraphics[width=\mywidth\textwidth]{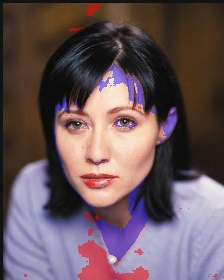} &
    \includegraphics[width=\mywidth\textwidth]{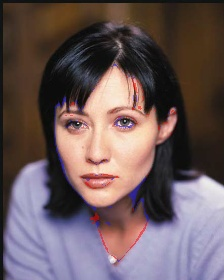} &
    \includegraphics[width=\mywidth\textwidth]{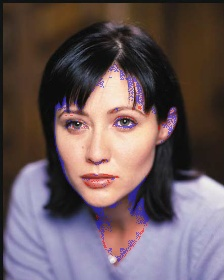} &
    \includegraphics[width=\mywidth\textwidth]{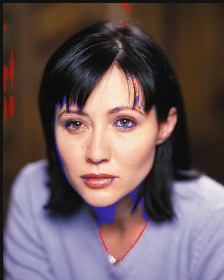} \\
& 0.9237	& 0.9216	& 0.7918	& \textbf{0.9611}	& 0.9242	& 0.9248\\

    \end{tabular}
    }

    \caption{Examples of skin segmentation performed using different techniques. False positives are indicated with red color and false negatives with blue color, and F-score is provided under each outcome (the best score for each example is boldfaced).}
    \label{fig:examples_good}
\end{figure}

\begin{figure}
\newcommand{\mywidth}{0.145}
    \centering
    \scriptsize
    \resizebox{\textwidth}{!}{
    \begin{tabular}{lcccccc}
    & \rgb & \nonskin & \gs & \ethree & \vote & Ens.-BC+S \\

    \raisebox{5mm}{(i)} &
     \includegraphics[width=\mywidth\textwidth]{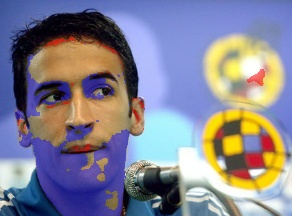} &
     \includegraphics[width=\mywidth\textwidth]{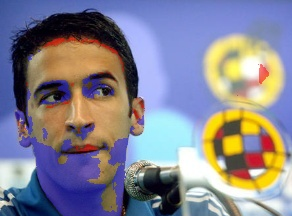} &
    \includegraphics[width=\mywidth\textwidth]{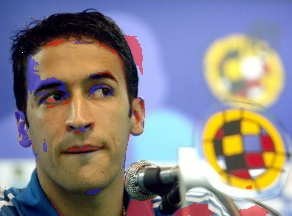} &
    \includegraphics[width=\mywidth\textwidth]{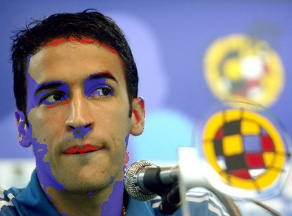} &
    \includegraphics[width=\mywidth\textwidth]{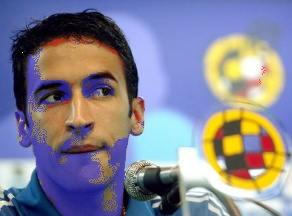} &
    \includegraphics[width=\mywidth\textwidth]{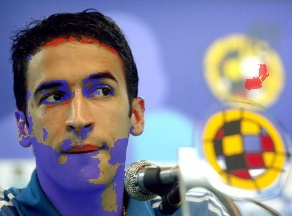} \\
& 0.6711	& 0.7357	& \textbf{0.8920}	& 0.8748	& 0.6895	& 0.7314\\


    \raisebox{5mm}{(ii)} &
    \includegraphics[width=\mywidth\textwidth]{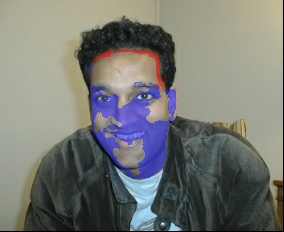} &
    \includegraphics[width=\mywidth\textwidth]{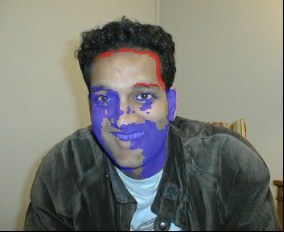} &
    \includegraphics[width=\mywidth\textwidth]{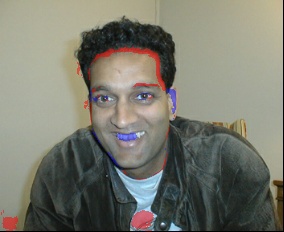} &
    \includegraphics[width=\mywidth\textwidth]{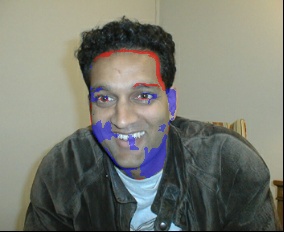} &
    \includegraphics[width=\mywidth\textwidth]{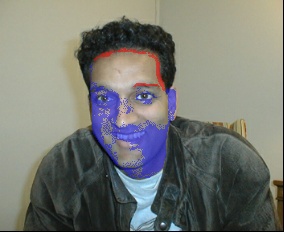} &
    \includegraphics[width=\mywidth\textwidth]{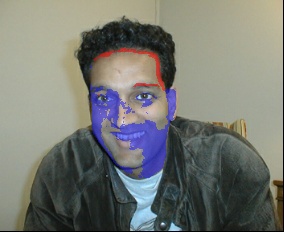} \\
& 0.6605	& 0.7240	& \textbf{0.9076}	& 0.8682	& 0.7187	& 0.7286\\

    \raisebox{5mm}{(iii)} &
     \includegraphics[width=\mywidth\textwidth]{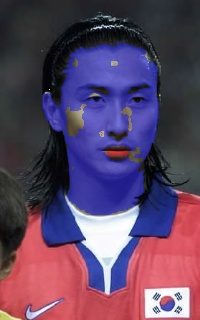} &
     \includegraphics[width=\mywidth\textwidth]{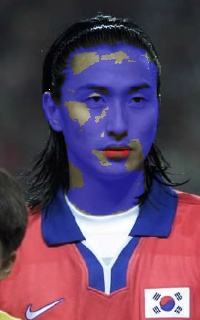} &
    \includegraphics[width=\mywidth\textwidth]{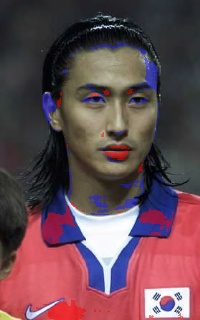} &
    \includegraphics[width=\mywidth\textwidth]{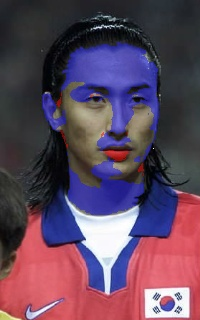} &
    \includegraphics[width=\mywidth\textwidth]{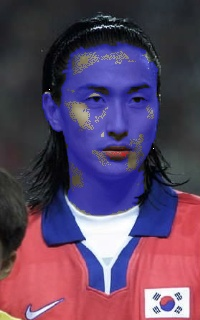} &
    \includegraphics[width=\mywidth\textwidth]{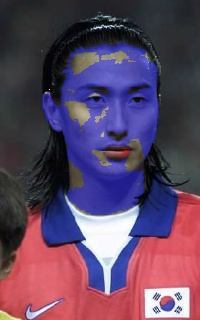} \\
& 0.1036	& 0.2565	& \textbf{0.8708}	& 0.5423	& 0.1979	& 0.2570\\

    \raisebox{5mm}{(iv)} &
     \includegraphics[width=\mywidth\textwidth]{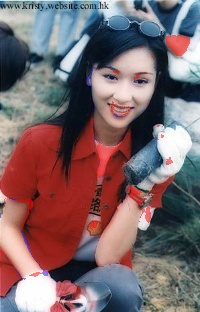} &
     \includegraphics[width=\mywidth\textwidth]{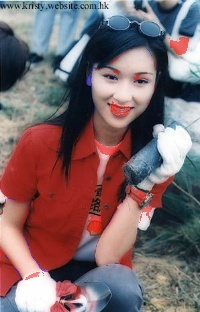} &
    \includegraphics[width=\mywidth\textwidth]{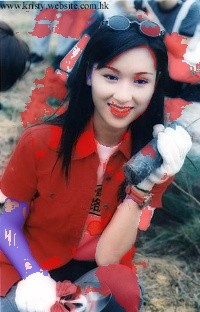} &
    \includegraphics[width=\mywidth\textwidth]{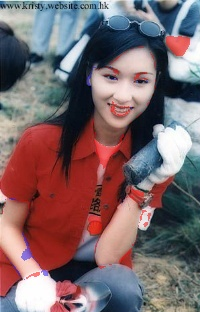} &
    \includegraphics[width=\mywidth\textwidth]{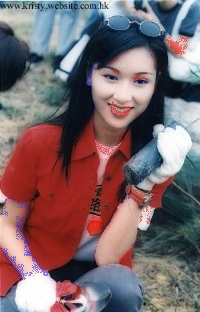} &
    \includegraphics[width=\mywidth\textwidth]{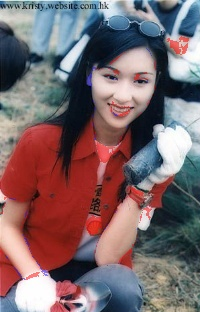} \\
& 0.9020	& \textbf{0.9180}	& 0.5315	& 0.8977	& 0.9111	& 0.9193\\

    \raisebox{5mm}{(v)} &
     \includegraphics[width=\mywidth\textwidth]{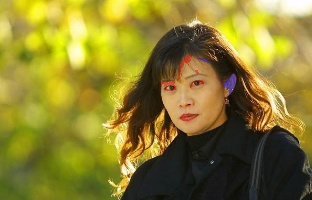} &
     \includegraphics[width=\mywidth\textwidth]{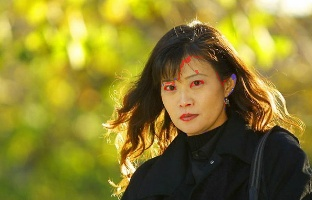} &
    \includegraphics[width=\mywidth\textwidth]{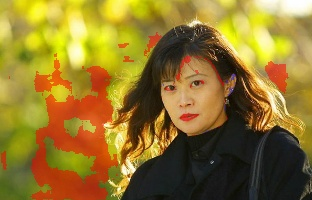} &
    \includegraphics[width=\mywidth\textwidth]{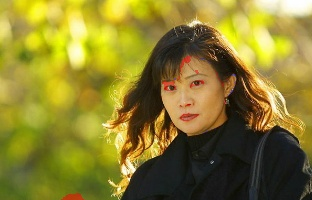} &
    \includegraphics[width=\mywidth\textwidth]{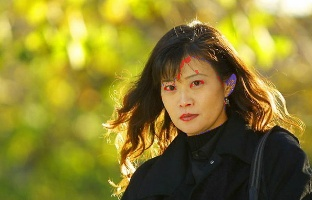} &
    \includegraphics[width=\mywidth\textwidth]{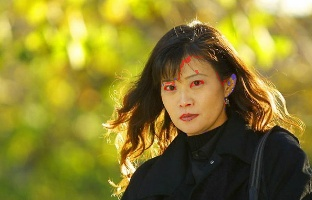} \\
& 0.9356	& \textbf{0.9570}	& 0.4547	& 0.9236	& 0.9527	& 0.9559\\

    \raisebox{5mm}{(vi)} &
     \includegraphics[width=\mywidth\textwidth]{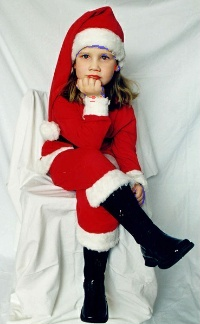} &
     \includegraphics[width=\mywidth\textwidth]{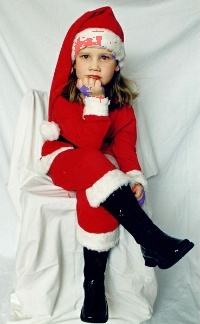} &
    \includegraphics[width=\mywidth\textwidth]{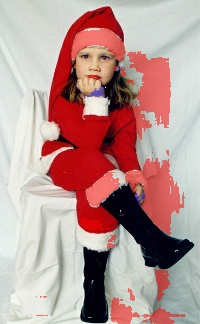} &
    \includegraphics[width=\mywidth\textwidth]{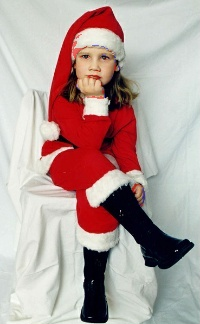} &
    \includegraphics[width=\mywidth\textwidth]{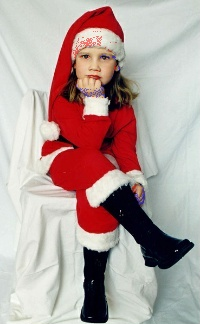} &
    \includegraphics[width=\mywidth\textwidth]{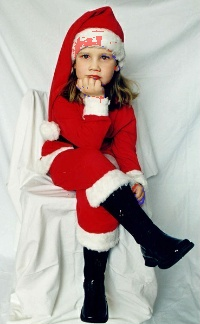} \\
& \textbf{0.9341}	& 0.8614	& 0.2742	& 0.9213	& 0.8820	& 0.8837\\

    \raisebox{5mm}{(vii)} &
     \includegraphics[width=\mywidth\textwidth]{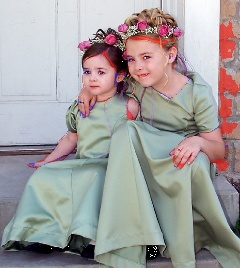} &
     \includegraphics[width=\mywidth\textwidth]{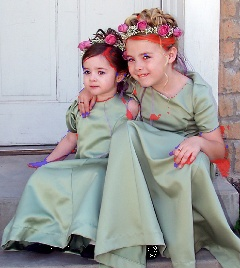} &
    \includegraphics[width=\mywidth\textwidth]{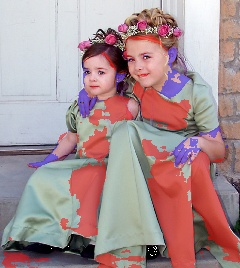} &
    \includegraphics[width=\mywidth\textwidth]{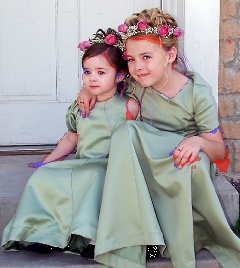} &
    \includegraphics[width=\mywidth\textwidth]{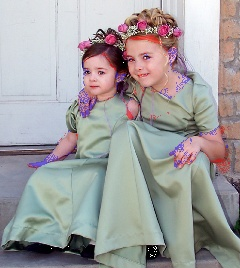} &
    \includegraphics[width=\mywidth\textwidth]{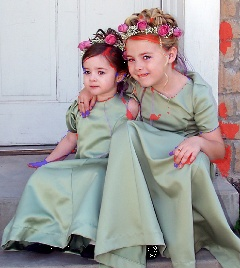} \\
& 0.8391	& 0.8991	& 0.3716	& \textbf{0.9077}	& 0.8832	& 0.8595\\

    \end{tabular}
    }
    \caption{Examples of skin segmentation for more challenging cases performed using different techniques. False positives are indicated with red color and false negatives with blue color, and F-score is provided under each outcome  (the best score for each example is boldfaced).}
    \label{fig:examples_difficult}
\end{figure}
In Figs.~\ref{fig:examples_good} and \ref{fig:examples_difficult}, we present examples of the segmentation outcome that allow us to analyze the results qualitatively. We showcase the results obtained using the vanilla \rgb\, model along with \nonskin\, and \gs\, base models, and we also include the outcomes retrieved using three ensembles of different kind: the Skinny-based \ethree, the voting-based \vote, and the two-branch \scombined. In Fig.~\ref{fig:examples_good}, both the RGB and grayscale base models render rather good results, and \ethree\, allows for improving the outcome in all the cases, reducing both false positives (red) and false negatives (blue). It is worth noting that for (a, b, d) the F-score is increased by around 0.1 compared with the base models, and it is also definitely more improved compared with the two remaining ensembles. In Fig.~\ref{fig:examples_difficult}, we present several more challenging cases, in which either the color models (i, ii, iii) or the \gs\, model (iv, v, vi, vii) fail to segment the skin regions accurately. Although the proposed \ethree\, surpasses all the base models only in (vii), it is clear that in the remaining cases it manages to effectively combine the base models, rendering the scores that are close to the best one, better than the voting and two-branch schemes. Even in the example (iii), where the color-based models fail completely in detecting the skin, \ethree\, manages to reduce the false negatives significantly (which does not happen for the two remaining ensembles). Also, false negatives are substantially reduced for (i) and (ii) relying on the outcome retrieved with \gs. In cases (vi) and (vii), \gs\, renders high false positive errors, which are successfully compensated by the ensemble models---as the remaining base models segment the skin regions mostly correctly here, the voting and two-branch schemes are also much better than \gs, however they are still outperformed with \ethree.

\section{Conclusions and Future Work} \label{sec:concl}

In this paper, we proposed a new approach towards creating ensembles for segmenting skin regions from color images. We benefit from learning a set of homogeneous base models (based on the Skinny CNN architecture~\cite{Tarasiewicz2020}), whose diversity is achieved by guiding the training to obtain models that are focused on different features. Throughout our experimental study, we demonstrated quantitatively and qualitatively that the models trained from grayscale and color images, including training data stratification relying on the BC, are complementary and therefore may be effectively combined relying on ensemble learning. Furthermore, the reported results indicate that the proposed ensembling scheme which extracts the contextual spatial features of the initial skin probability maps using a second-level CNN outperforms the pixel-wise voting that is already a well-established approach for semantic segmentation~\cite{Nanni2023}.

The reported results are encouraging and in our future work we want to further extend the proposed approach. Our study was limited to creating two-level CNN ensembles---we expect that multi-level structures, possibly composed of lighter models, may improve the segmentation outcome even further, while making it possible to control the balance between the processing time and segmentation quality. While ensembling neural networks is quite convenient for creating resource-frugal solutions~\cite{Liang2019}, in our approach, the segmentation outcome can be gradually refined when passed through subsequent levels of the sequentially-connected CNNs, so by choosing arbitrarily the number of processing levels employed during segmentation, the time vs. quality trade-off can be managed effectively. Also, it may be worth considering to fine-tune the connected CNNs, thus resulting in a multi-branch architecture. Last, but not least, we believe the developed approach can also be applied to other semantic segmentation tasks~\cite{HaoZhou2020}, in particular those that can benefit from heterogeneous image fusion.




%
%
%

\section*{Acknowledgements}

This work was supported by the National Science Centre, Poland, under Research Grant 2022/47/B/ST6/03009.

\end{document}